\documentclass[11pt,a4paper]{article}

\usepackage[utf8]{inputenc}
\usepackage[T1]{fontenc}
\usepackage[english]{babel}
\usepackage{lmodern}
\usepackage{microtype}
\usepackage{geometry}
\usepackage{booktabs}
\usepackage{tabularx}
\usepackage{array}
\usepackage{amsmath}
\usepackage{graphicx}
\graphicspath{{figures/}}
\usepackage{hyperref}
\usepackage{url}
\usepackage{parskip}

\hypersetup{
  colorlinks=true,
  linkcolor=blue,
  citecolor=blue,
  urlcolor=blue
}

\newcolumntype{Y}{>{\raggedright\arraybackslash}X}

\title{What makes prompts a graph: necessary and sufficient conditions for prompt graph engineering}
\author{Sandeco Macedo\\
\textit{Federal Institute of Goiás}\\
\texttt{sanderson.macedo@ifg.edu.br}}
\date{\today}

\begin{document}

\maketitle

\begin{abstract}
Prompts stopped being isolated strings some time ago. In real systems, one model call feeds another, retrieval interleaves with generation, routers branch, and aggregators merge parallel results. Practice converged on a single structure to hold this together: the graph. Frameworks such as LangGraph, DSPy, and Prompt Flow expose it openly, and research systems already optimize it automatically. The vocabulary, however, lags behind. \textit{Graph} names, variously, a reasoning topology inside one sampling strategy, a multi-agent conversation, or an orchestration artifact, while \textit{prompt engineering} still evokes writing one good string. What is missing is a reference definition treating prompts as nodes of an explicit, executable, improvable graph. We build that definition through conceptual analysis over sources with persistent identifiers, complemented by primary grey literature. We reconstruct the genealogy of the idea, from dataflow graphs and build systems, through prompt chaining and the thought topologies (chain, tree, graph), to graphs compiled and optimized as artifacts. We then propose a constitutive definition of prompt graph engineering, state its four conditions (explicit structure, separation between structure and prompt content, executable semantics, and the graph as a first-class engineering artifact), and operationalize them as an inclusion and exclusion test. We draw the boundary against six neighboring concepts and apply the test to six real systems (LangGraph, DSPy, Prompt Flow, AutoGen, CrewAI, and Claude Code subagents); it includes and excludes consistently. We close with a research agenda organized along four design tension axes. The contribution is an operational definition and a shared vocabulary for a practice that industry already exercises daily without naming precisely.
\end{abstract}

\noindent\textbf{Keywords:} prompt graph engineering; prompt engineering; large language models; LLM pipelines; agentic workflows; graph of thoughts; operational definition.

\section{Introduction}

Ask an engineer to show you the prompt of a modern language model application and the honest answer is that there is no such prompt. There is a retrieval step that fetches context, a first call that plans, a router that decides which specialist handles the case, three calls that run in parallel, an aggregator that votes on their outputs, and a verifier that refuses to pass along an answer that fails its checks. The unit of engineering is no longer a string. It is a structure that connects many strings, and that structure has a mathematical name that practitioners already use without ceremony: a graph, with prompts on the nodes and dependencies on the edges.

The path to that structure was fast. Large language models first showed that a single well-crafted prompt could steer a general model to a task \cite{Brown2020FewShot}, and instruction tuning made that steering reliable enough to build on \cite{Ouyang2022InstructGPT,Zhao2023SurveyLLM}. A craft grew around the single string, with its patterns, catalogs, and surveys \cite{Liu2023PretrainPrompt,Sahoo2024PromptSurvey,Schulhoff2024PromptReport}. But the single string hit its ceiling early. Complex tasks demanded decomposition, and decomposition produced multiplicity: chains of model calls curated by humans \cite{Wu2022AIChains}, chains of intermediate reasoning inside one generation \cite{Wei2022CoT}, trees that branch and backtrack \cite{Yao2023ToT}, and finally graphs, where partial results merge, loop, and feed one another \cite{Besta2024GoT}. In parallel, an engineering lineage turned those structures into programs: declarative pipelines that compile \cite{Khattab2023DSPy}, function-call DAGs that schedule in parallel \cite{Kim2023LLMCompiler}, agent workflows represented and even searched as graphs \cite{Zhuge2024GPTSwarm,Zhang2024AFlow}. Products followed the same drift; LangGraph names the graph in its own name, and Prompt Flow renders it on screen.\footnote{LangGraph documentation: \url{https://langchain-ai.github.io/langgraph/}. Microsoft Prompt Flow documentation: \url{https://microsoft.github.io/promptflow/}. Accessed July 2026.}

The structure converged; the vocabulary did not. \textit{Graph} names at least three different things in current usage. In graph-of-thoughts it names a topology of model-generated thoughts inside a single problem-solving episode \cite{Besta2024GoT}. In multi-agent systems it names, often only implicitly, whatever shape a free conversation among agents happens to trace \cite{Wang2024SurveyAgents}. In orchestration frameworks it names an authored artifact, drawn or coded by an engineer, that a runtime executes. Meanwhile \textit{prompt engineering} still evokes the craft of one good string \cite{Sahoo2024PromptSurvey}. The closest existing name, \textit{flow engineering}, coined for the code-generation flow of AlphaCodium, points at the practice but stops at pointing: it names the shift from the single prompt to a structured flow of calls without saying which conditions a flow must meet, and a flow, unlike a graph, does not commit to explicit routing, merging, or artifact status \cite{Ridnik2024AlphaCodium}. What is missing, then, is not a word but a constitutive definition of the discipline that sits between prompt craft and agent architecture: the engineering of the structure itself, of how prompt-bearing nodes compose, route, and execute. The polysemy is not cosmetic. It prevents us from saying which systems share a design, from comparing a compiled DSPy pipeline with a hand-drawn flow on equal terms, and from accumulating knowledge about when structure, rather than prompt wording, is what improves a system \cite{Gao2023RAGSurvey}.

The practical consequence is familiar to anyone who has debugged one of these systems. When the flow lives inside an opaque script, or inside the improvised turns of a conversation among agents, there is nothing to inspect when it misbehaves, nothing to type-check before running it, nothing to version when it changes, and nothing for an optimizer to hold on to. The lineage of automatic optimizers makes the point sharply: a structure can only be tuned if it exists as an object \cite{Yuksekgonul2024TextGrad,Zhang2024AFlow}. Naming the discipline that produces such objects is the purpose of this article.

This article delivers a reference definition of \textit{prompt graph engineering}. The central question is direct: what conditions are necessary and sufficient for a practice, or a system, to count as engineering prompts as a graph, and how do we turn that into an instrument that discriminates cases? We organize five contributions around five research questions. \textbf{RQ1 (genealogy):} where does the idea of structuring prompts as a graph come from, and how did its sense migrate from reasoning topology to engineering artifact? \textbf{RQ2 (constitutive definition):} what conditions are necessary and sufficient for prompt graph engineering? \textbf{RQ3 (boundary):} how does the definition distinguish the concept from classic prompt engineering, thought topologies, agent orchestration, prompt programming, RAG pipelines, and classic workflow engines? \textbf{RQ4 (application):} how do real systems instantiate the constitutive conditions? \textbf{RQ5 (agenda):} which design axes remain open?

The contributions are, in order: a genealogy that reconstructs the migration of the graph from mathematical model of computation to engineering artifact for prompts (Section~\ref{sec:genealogia}); a constitutive definition operationalized as an inclusion and exclusion test (Section~\ref{sec:definicao}); a boundary delimitation against six neighboring concepts (Section~\ref{sec:fronteira}); the application of the definition to six real systems, showing that it classifies consistently (Section~\ref{sec:aplicacao}); and a research agenda organized by tension axes (Section~\ref{sec:agenda}). Section~\ref{sec:relacionados} positions the article against the literature.

Positioning. This article is deliberately definitional. It does not propose a new framework, does not benchmark systems, and does not survey the prompt engineering literature, which already has its surveys \cite{Schulhoff2024PromptReport}. Its product is conceptual: a shared vocabulary and a reproducible test of membership in the concept. A note on method: this is a work of conceptual clarification; each source with a verified DOI enters as a formal citation, while product documentation, where much of the practice actually lives, enters as footnotes with URLs.

\section{Related Work}
\label{sec:relacionados}

Five research lines surround the concept without defining it, and three smaller neighborhoods, the interface antecedents, the terminological plane, and the definitional method, complete the picture. The prompt engineering surveys systematize techniques for the individual prompt and its variations, from the pre-train, prompt, and predict paradigm to pattern catalogs and broad technique taxonomies \cite{Liu2023PretrainPrompt,White2023PromptPatterns,Sahoo2024PromptSurvey,Schulhoff2024PromptReport}; their unit of analysis is the prompt, not the structure that connects prompts. The taxonomies of thought topologies compare chain, tree, and graph as shapes of intermediate reasoning \cite{Besta2024Demystifying,Xia2024ChainOfX,Ding2023EoT}; they classify topologies inside a solving episode and stop short of the graph as an authored engineering object. The agent surveys describe architectures, components, and design patterns of LLM-based agents and multi-agent systems \cite{Wang2024SurveyAgents,Xi2023RisePotential,Guo2024MultiAgentSurvey,Masterman2024Landscape,Liu2024DesignPatterns}; the flow among agents appears as a byproduct of the architecture, rarely as the object itself. The programming lineage gives prompts formal structure, from probabilistic cascades over model calls \cite{Dohan2022Cascades}, to query and programming languages \cite{BeurerKellner2023LMQL,Vaziri2024PDL}, to declarative pipelines that compile \cite{Khattab2022DSP,Khattab2023DSPy}, to runtimes that schedule structured programs \cite{Zheng2023SGLang,Kim2023LLMCompiler}; it builds instances of the concept without naming or delimiting it. And the optimization lineage treats prompts, and lately whole workflows, as search spaces \cite{Zhou2023APE,Fernando2023Promptbreeder,Yuksekgonul2024TextGrad,Hu2024ADAS,Zhang2024AFlow,Zhuge2024GPTSwarm}; it presupposes exactly the artifact this article defines, since only an explicit graph can be searched. The closest antecedents come from human-computer interaction, where chaining model calls and editing the chain visually were proposed early \cite{Wu2022AIChains,Wu2022PromptChainer}; those works demonstrated the value of the artifact but framed it as an interface technique, not as a discipline with membership conditions. On the terminological side, \textit{flow engineering}, coined alongside AlphaCodium to name iterative, structured code-generation flows \cite{Ridnik2024AlphaCodium}, baptizes the practice without delimiting it, offering neither conditions of membership nor a boundary against neighboring concepts; and \textit{compound AI systems} names the broad class of systems that tackle AI tasks with multiple interacting components, model calls, retrievers, and tools, delimiting the class by component count rather than by structure, and demanding neither an explicit graph, nor separation of structure and content, nor artifact status.\footnote{Zaharia et al., \textit{The Shift from Models to Compound AI Systems}, BAIR Blog, February 2024: \url{https://bair.berkeley.edu/blog/2024/02/18/compound-ai-systems/}. Accessed July 2026.} In method, the closest neighbors of this article are two earlier definitional studies by the same author, on adjacent layers of the same stack: one gives necessary and sufficient conditions for the \textit{agent harness}, the runtime that turns an LLM into an agent \cite{Macedo2026Harness}, and the other defines \textit{loop engineering}, the design of the external loop that drives a harness in place of step-by-step prompting, placed in the progression from prompt to context, harness, and loop \cite{Macedo2026LoopEng}. This article defines the layer that organizes the prompts themselves: where the loop drives the agent from the outside, the graph structures the composition of calls on the inside. What no line delivers is the constitutive definition: which conditions make a system an instance of prompt graph engineering, and which neighbors fail which condition. For the definitional method itself we follow the classic criteria for designing shared conceptualizations \cite{Gruber1995Ontology}.

\section{Genealogy (RQ1)}
\label{sec:genealogia}

The idea that computation should be drawn as a graph is half a century older than the prompt. Dataflow models represented a program as nodes that fire when their inputs arrive, with edges carrying data between them \cite{Dennis1974Dataflow,DavisKeller1982Dataflow}. The same shape reappeared wherever engineers needed to coordinate steps they did not want to weld together: Make expressed a build as a dependency graph among targets \cite{Feldman1979Make}, and scientific workflow systems scaled the pattern to entire experiments, with the graph as the unit that is shared, audited, and rerun \cite{Deelman2009Workflows}. Three lessons from that tradition matter here. The graph separates orchestration from computation. The graph makes dependencies explicit, so independent work can run in parallel. And the graph is an artifact: it can be checked before running and inspected after failing.

The prompt entered this story innocent of all that. A prompt was a string handed to a model, and the early craft consisted of making that one string work: few-shot examples in context \cite{Brown2020FewShot}, instructions that models learned to follow \cite{Ouyang2022InstructGPT}, even the bare magic words that elicit step-by-step reasoning \cite{Kojima2022ZeroShot}. The craft matured into catalogs and surveys \cite{Liu2023PretrainPrompt,White2023PromptPatterns}, but its unit never stopped being the individual string. The first crack in that unit came from decomposition. Least-to-most prompting solved a hard problem by chaining the solutions of its subproblems \cite{Zhou2023LeastToMost}, and decomposed prompting made the move architectural: a library of subtask handlers that a decomposer prompt invokes, explicitly modular \cite{Khot2023Decomposed}. Once a task spans multiple calls, something must say which call feeds which. Structure had arrived, even if nobody drew it yet.

Two lineages then took the structure in different directions. The first kept it inside the model's reasoning. Chain-of-thought made the model produce a linear sequence of intermediate steps within a single generation \cite{Wei2022CoT}; self-consistency sampled many chains in parallel and voted \cite{Wang2023SelfConsistency}; tree-of-thoughts turned generation into search over a branching space of partial thoughts \cite{Yao2023ToT}; and graph-of-thoughts generalized the shape to arbitrary graphs where thoughts merge, refine, and loop \cite{Besta2024GoT}, with variants that condition generation on graph-structured context \cite{Yao2023GoTReasoning} and skeleton-first strategies that parallelize expansion \cite{Ning2023SoT}. The community itself began classifying these shapes as topologies of reasoning \cite{Besta2024Demystifying,Xia2024ChainOfX,Ding2023EoT}. In this lineage the graph describes what the model thinks. Its nodes are thoughts the model generated; its shape is chosen by a search strategy, per problem instance.

The second lineage pulled the structure out of the model and into the engineer's hands. AI Chains showed that humans compose chains of model calls, each call doing one thing, and that the chain itself improves outcomes and controllability \cite{Wu2022AIChains}; PromptChainer gave that chain a visual editor, nodes on a canvas, edges dragged by hand \cite{Wu2022PromptChainer}. Language model cascades framed multi-call compositions as probabilistic programs, giving the practice formal semantics \cite{Dohan2022Cascades}. Offloading turned nodes heterogeneous: a node could run code rather than sample a model \cite{Gao2023PAL,Chen2023PoT}. Then the programming lineage industrialized the artifact. Demonstrate-search-predict composed retrieval and generation into declarative pipelines \cite{Khattab2022DSP}; DSPy separated the program's structure from the prompt text of its modules and compiled the latter against the former \cite{Khattab2023DSPy}; LMQL and PDL made calls and constraints part of a language \cite{BeurerKellner2023LMQL,Vaziri2024PDL}; SGLang gave structured programs a runtime \cite{Zheng2023SGLang}; and LLMCompiler scheduled function calls as a dependency DAG, recovering the parallelism argument of dataflow almost verbatim \cite{Kim2023LLMCompiler}.

The agentic wave completed the migration. Multi-agent systems composed role-played model instances into conversations \cite{Li2023CAMEL,Wu2023AutoGen}, standardized operating procedures \cite{Hong2023MetaGPT}, chat chains for software development \cite{Qian2023ChatDev}, and explicit state machines \cite{Wu2024StateFlow}, with flows proposed as the general abstraction for structured interaction \cite{Josifoski2023Flows}. And then the artifact became not only explicit but optimizable: GPTSwarm represented agent societies literally as computational graphs whose nodes and edges are optimized \cite{Zhuge2024GPTSwarm}; ADAS searched the space of agent designs in code \cite{Hu2024ADAS}; AFlow searched workflow graphs by Monte Carlo tree search \cite{Zhang2024AFlow}. Products consolidated the practice: LangGraph exposes a state graph as its central API, Prompt Flow renders a DAG of prompt and tool nodes, and visual builders put the canvas in front of non-programmers.\footnote{LangGraph: \url{https://langchain-ai.github.io/langgraph/}, announced in January 2024: \url{https://blog.langchain.com/langgraph/}. Prompt Flow: \url{https://microsoft.github.io/promptflow/}, first public release in 2023: \url{https://github.com/microsoft/promptflow}. Flowise: \url{https://flowiseai.com/}. Similar canvases predate LLM pipelines in other domains, for example ComfyUI for image generation: \url{https://github.com/comfyanonymous/ComfyUI}. Accessed July 2026.}

\begin{figure}[h!]
    \centering
    \includegraphics[width=0.98\linewidth]{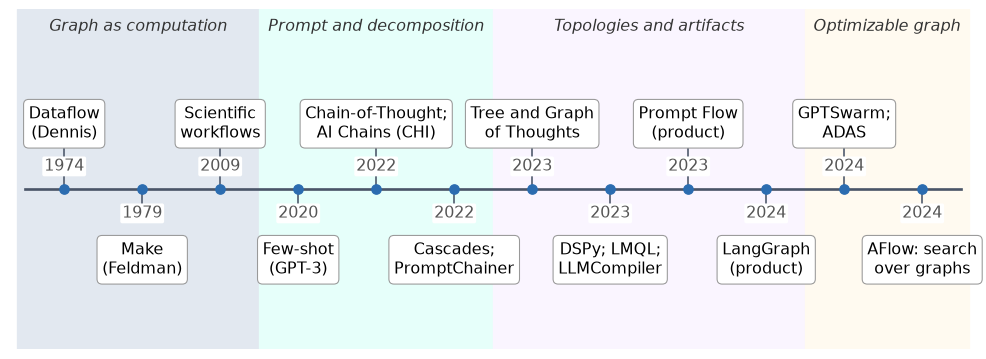}
    \caption{Genealogy of the prompt graph. The graph as model of computation (dataflow, build systems, scientific workflows) meets the prompt (few-shot, instruction following), passes through decomposition and chaining, splits into reasoning topologies (chain, tree, graph of thoughts) and engineering artifacts (cascades, DSP/DSPy, LLMCompiler, agent workflow graphs), and converges on graphs that are compiled and optimized as first-class objects. The background bands mark periods, not categories: each event sits in its date's band, whatever its lineage.}
    \label{fig:timeline}
\end{figure}

\begin{figure}[h!]
    \centering
    \includegraphics[width=0.98\linewidth]{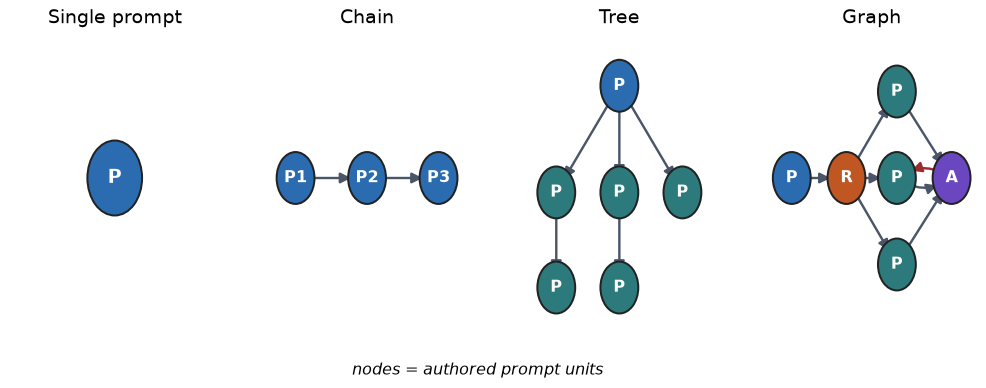}
    \caption{From string to graph. The four shapes of prompt-mediated computation: the single prompt; the chain, where each call feeds the next; the tree, where branches explore alternatives; and the graph, where routing, parallel branches, aggregation, and cycles compose freely: node R routes, node A aggregates, and the pair of edges between the central node and A (forth and back) forms a refinement cycle. In prompt graph engineering the nodes are authored prompt units, not model-generated thoughts.}
    \label{fig:topologies}
\end{figure}

Figure~\ref{fig:timeline} condenses the migration; Figure~\ref{fig:topologies} shows the shapes it passed through. Two histories must be kept apart here, because their chronologies differ. The structure came first through chains: the engineering lineage was already composing authored calls in 2021 and 2022 \cite{Wu2022AIChains,Dohan2022Cascades}, before any thought topology carried the word \textit{graph}. The word came later and from the other lineage: \textit{graph} entered the shared vocabulary through graph-of-thoughts in 2023 \cite{Besta2024GoT} and was claimed by the engineering side within months, in product names and in research systems \cite{Zhuge2024GPTSwarm}. So the migration this genealogy establishes is of the word's dominant sense, not of the practice: the practice grew continuously out of chaining, while the word \textit{graph} moved from mathematics to reasoning topology to engineering artifact. In that final sense the object changed owners, from the search strategy to the engineer, and that is the sense, the graph as the unit of engineering for prompt-mediated computation, that this article defines.

\section{Constitutive Definition (RQ2)}
\label{sec:definicao}

The genealogy tells us where the structure comes from. What remains is to say what the discipline is, and precisely. We define prompt graph engineering by conditions, not by examples, and we turn the definition into a test that includes and excludes concrete cases: this is our answer to RQ2. A good constitutive definition enumerates what is necessary, drops what is incidental, and separates the concept from its neighbors \cite{Gruber1995Ontology}.

We propose the following reference definition.

\begin{quote}
\textit{Prompt graph engineering is the discipline that represents, composes, and executes prompt-mediated language model computation as an explicit graph, in which: (i) the nodes are authored units of computation, prompt-parameterized model invocations or deterministic transforms, and the edges are data or control dependencies among them; (ii) the structure of the graph is separated from the prompt content of its nodes, so that either can change without rewriting the other; (iii) the graph has executable semantics, a runtime that schedules nodes, routes outputs, and manages shared state, including branching, parallelism, and cycles; and (iv) the graph is a first-class engineering artifact, an object that can be inspected, versioned, validated, and optimized independently of any particular execution.}
\end{quote}

A practice, or a system, belongs to prompt graph engineering if and only if it instantiates the four elements above, which we will call G1 (explicit structure), G2 (separation of structure and content), G3 (executable semantics), and G4 (first-class artifact status). The definition states conditions, not cases, and each condition earns its place by lineage. Explicit structure with firing nodes and dependency edges is the dataflow inheritance \cite{Dennis1974Dataflow,DavisKeller1982Dataflow}. Separation between structure and content is what made chains editable in the first HCI systems \cite{Wu2022AIChains,Wu2022PromptChainer} and what DSPy radicalized by compiling prompt text against a fixed program structure \cite{Khattab2023DSPy}. Executable semantics is what distinguishes an architecture diagram from a program, whether the semantics is probabilistic \cite{Dohan2022Cascades}, scheduled as a parallel DAG \cite{Kim2023LLMCompiler}, or interpreted over shared state by a product runtime.\footnote{LangGraph's StateGraph is the clearest product example: nodes update a shared typed state, edges and conditional edges route among them: \url{https://langchain-ai.github.io/langgraph/}. Accessed July 2026.} And first-class artifact status is what the optimization lineage presupposes: a graph can only be searched, differentiated through, or refactored if it exists as an object \cite{Zhuge2024GPTSwarm,Zhang2024AFlow,Yuksekgonul2024TextGrad}. A note on the order of derivation, to keep the definition honest: the four conditions were fixed from these lineages, dataflow, HCI chaining, programming, and optimization, before any present-day framework was examined, and Section~\ref{sec:aplicacao} then applies them to systems chosen to span the practice. The test is not tailored to bless the systems that inspired the genealogy; one indication is that it excludes a popular contemporary system outright.

\begin{figure}[h!]
    \centering
    \includegraphics[width=0.96\linewidth]{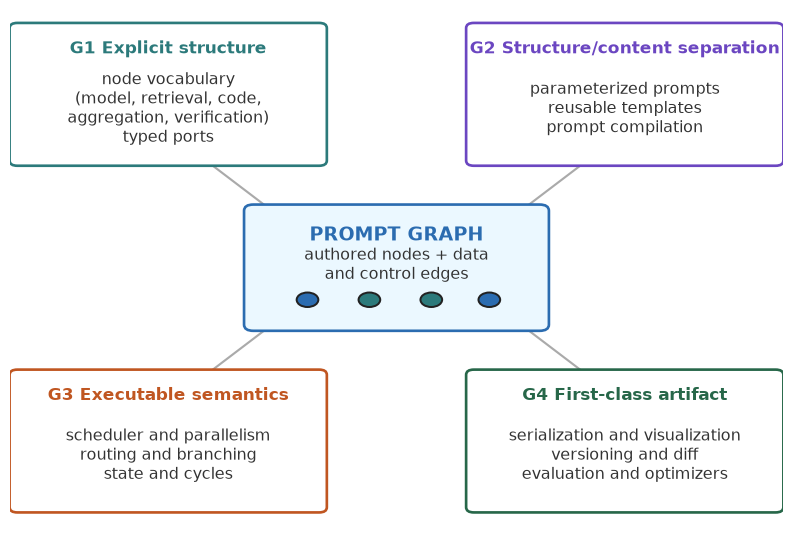}
    \caption{Anatomy of prompt graph engineering. At the center, the graph: authored prompt and transform nodes connected by data and control edges. Around it, what the four conditions demand: node vocabulary and typed ports (G1), parameterization that separates structure from prompt content (G2), scheduler, state, and routing (G3), and artifact services such as serialization, visualization, evaluation, and optimization (G4).}
    \label{fig:anatomy}
\end{figure}

Figure~\ref{fig:anatomy} opens the concept into components. At the center, the graph itself: prompt nodes, transform nodes, and the edges among them. Around it, the machinery the four conditions demand: a node vocabulary (model call, retrieval, code execution, aggregation, verification \cite{Lewis2020RAG,Gao2023PAL,Wang2023SelfConsistency,Madaan2023SelfRefine}), typed ports that make dependencies checkable, a scheduler that exploits the parallelism the edges reveal \cite{Kim2023LLMCompiler}, state and routing for cycles and feedback loops \cite{Shinn2023Reflexion,Wu2024StateFlow}, and the artifact services: serialization, visualization, versioning, evaluation, and optimizers that improve prompts given their position in the structure \cite{Khattab2023DSPy,Yuksekgonul2024TextGrad}. Not every system ships every component. The four conditions, though, are the core; without them there is no prompt graph engineering.

Each condition is necessary, and removing one at a time shows why. Remove explicit structure (G1) and the flow retreats into code paths or conversation turns; the system may work, but nothing enumerates its steps, and the practice collapses back into script writing, exactly the opacity that chaining research set out to cure \cite{Wu2022AIChains}. Remove the separation of structure and content (G2) and every structural change forces prompt rewrites and vice versa; nodes cannot be reused, and no optimizer can hold the structure fixed while tuning the text, the very operation DSPy defines \cite{Khattab2023DSPy}. Remove executable semantics (G3) and the graph is a picture; architecture diagrams of agent systems abound in surveys \cite{Masterman2024Landscape}, and none of them runs. Remove artifact status (G4) and the graph exists only at runtime, as a trace; it cannot be versioned, statically checked, or searched, which forfeits the entire optimization lineage \cite{Hu2024ADAS,Zhang2024AFlow}. The four together are sufficient: a system that satisfies G1 through G4 already exhibits everything the concept was coined to capture, structure you can see, edit, run, and improve. There is no fifth condition to add; observability, caching, cost control, and human-in-the-loop nodes are specializations of the four, not new elements.

What is NOT required also has to be said, and the definition stays lean because it leaves the incidental out. Prompt graph engineering does not require a visual editor: a graph defined in code, as in DSPy modules or LangGraph builders, is exactly as explicit as one drawn on a canvas \cite{Khattab2023DSPy}. It does not require acyclicity: cycles with exit conditions are constitutive of reflection and retry loops \cite{Shinn2023Reflexion,Madaan2023SelfRefine}, and the definition demands executable semantics for them, not their absence. It does not require multiple models, a specific framework, or agents: a graph whose nodes all call one model is still a prompt graph, and agenthood is a property some nodes may have, not a condition on the graph. And it does not require automatic optimization: G4 demands that the graph be optimizable in principle, an object something could improve; hand-tuning a versioned graph satisfies it.

The definition becomes an instrument once we turn it into a decision procedure. Given a candidate practice or system, ask, in order: (T1) is there an explicit representation in which prompt-bearing computation units are nodes and their data or control dependencies are edges, enumerable without running the system? (T2) can the structure be changed without rewriting the prompt content of the nodes, and the content without changing the structure? (T3) does a runtime execute that representation, scheduling nodes, routing outputs, and managing state? (T4) is the graph available as an object beyond any single run, one that can be inspected, versioned, validated, or optimized? A candidate belongs to the concept if it answers \textit{yes} to T1 through T4. Fail any one and it falls into a neighboring category; Section~\ref{sec:fronteira} shows which. Table~\ref{tab:teste} summarizes the test.

\begin{table}[h!]
\centering
\small
\begin{tabularx}{\textwidth}{|c|Y|Y|}
\hline
\textbf{Test} & \textbf{Question} & \textbf{If the answer is \textit{no}} \\
\hline
T1 & Are prompt units nodes, and dependencies edges, in an explicit, enumerable representation? & Monolithic prompt, opaque script, or emergent conversation \\
\hline
T2 & Can structure and prompt content change independently of each other? & Welded chain, or nodes authored by the model rather than the engineer; structure and content cannot vary independently \\
\hline
T3 & Does a runtime execute the graph, with scheduling, routing, and state? & Architecture diagram or documentation drawing; it does not run \\
\hline
T4 & Is the graph an object beyond a single run, inspectable and improvable? & Ephemeral trace; nothing to version, check, or optimize \\
\hline
\end{tabularx}
\caption{Inclusion and exclusion test derived from the constitutive definition. A practice or system belongs to prompt graph engineering if it answers \textit{yes} to T1 through T4; each failure indicates the neighboring category into which the candidate falls.}
\label{tab:teste}
\end{table}

Two caveats keep the test from being read naively. The first concerns thresholds. T1 asks for more than plurality: two calls glued by string concatenation inside a function body are a chain in spirit but not an explicit representation; the criterion is that nodes and edges can be enumerated by inspection, by an API, or by serialization, without executing the system. T3 asks for more than sequencing: a runtime satisfies it when the graph, not hand-written control flow around it, determines what runs next; and dynamic graphs satisfy it as long as the construction rules are themselves explicit, the position taken by systems that build the DAG on the fly \cite{Kim2023LLMCompiler}. T4 has a verifiable criterion: some tool or process other than the executor must be able to consume the graph, a type checker, a visualizer, a diff, an optimizer; if the only consumer of the structure is the run itself, T4 fails. The second caveat: membership is binary, quality is gradual. A three-node retrieve, generate, verify graph serialized in YAML passes T1 through T4 and is a legitimate, if modest, instance of the concept. What separates it from a compiled DSPy program with an optimizer in the loop is not membership but maturity, above all in G4. The test decides \textit{whether} a practice is prompt graph engineering. The anatomy qualifiers measure \textit{how} developed it is. Conflating those two questions produces most of the terminological noise this article addresses.

\section{Boundary (RQ3)}
\label{sec:fronteira}

A definition proves itself at the edges. This section walks the boundary of prompt graph engineering against six neighboring concepts, and for each neighbor names the condition it fails. The neighbors are not adversaries; most are ingredients or ancestors of the concept. The point is that the test discriminates.

Classic prompt engineering is the nearest neighbor and the easiest to separate. Its craft is intra-node: wording, examples, format, patterns \cite{White2023PromptPatterns,Sahoo2024PromptSurvey,Schulhoff2024PromptReport}. It fails T1: there is one prompt, so there is no structure to represent. The relation between the two disciplines is containment of scope, not competition: every node of a prompt graph deserves good prompt engineering, and techniques cataloged at the prompt level parameterize nodes at the graph level. The graph does not make wording irrelevant; it makes wording local.

The thought topologies, chain, tree, and graph of thoughts, are the neighbor most often conflated with the concept, because they share the word \textit{graph}. Chain-of-thought fails T1 outright: its chain is a rhetorical structure inside one generated text, not a structure of invocations \cite{Wei2022CoT,Kojima2022ZeroShot}. Tree and graph of thoughts do orchestrate many invocations over an explicit topology maintained by a controller \cite{Yao2023ToT,Besta2024GoT}, and graph-of-thoughts even reifies a graph of operations, so T1 and T3 arguably pass. The failure is T2, and it is diagnostic: the nodes are thoughts the model generated, not prompt units an engineer authored; structure and content cannot vary independently because the content \textit{is} the model's output and the structure is the search schedule that produced it. The same holds for the broader X-of-thought family and its taxonomies \cite{Yao2023GoTReasoning,Ning2023SoT,Xia2024ChainOfX,Ding2023EoT,Besta2024Demystifying}: they are inference strategies, describable as graphs, that can be wrapped \textit{inside} a node of a prompt graph. The boundary line is authorship: in prompt graph engineering the engineer owns the nodes; in thought topologies the model does.

Agent orchestration overlaps the concept without coinciding with it. Multi-agent systems in free conversation, role-play societies, and group chats \cite{Li2023CAMEL,Wu2023AutoGen,Qian2023ChatDev} fail T1: the interaction shape emerges turn by turn, and no representation enumerates it beforehand; the surveys themselves describe these flows as emergent \cite{Wang2024SurveyAgents,Guo2024MultiAgentSurvey,Xi2023RisePotential}. Orchestration crosses the boundary exactly when the flow is reified: MetaGPT's standardized procedures \cite{Hong2023MetaGPT}, StateFlow's state machines \cite{Wu2024StateFlow}, and flows as a compositional abstraction \cite{Josifoski2023Flows} all pass T1 and T3, and pass T2 and T4 to the degree that the flow is authored and persists as an object. The granularity also differs: agent orchestration composes agents, units with goals, memory, and tools; prompt graph engineering composes computation at the finer grain of prompt-parameterized invocations. An agent can be a node, but the concept does not require nodes that thick.

Prompt programming, the lineage of LMQL, PDL, DSP, and DSPy, is not a neighbor to exclude but the code-shaped form of the concept, and drawing this boundary carefully matters because it shows the definition is about structure, not syntax. A DSPy program passes all four tests: modules are nodes (T1), signatures separate structure from prompt text, which the compiler generates (T2, radicalized), execution interprets the module composition (T3), and the program is exactly the artifact optimizers consume (T4) \cite{Khattab2022DSP,Khattab2023DSPy}. Language-level systems sit at the boundary's inner edge: LMQL structures constraints and control \textit{within} what is largely a single query's scope \cite{BeurerKellner2023LMQL}, and PDL makes call composition declarative \cite{Vaziri2024PDL}; they pass when a program composes multiple prompt-bearing units, and fail T1 when it structures only one. SGLang and LLMCompiler supply T3 machinery, runtimes and schedulers, for whoever authors the structure \cite{Zheng2023SGLang,Kim2023LLMCompiler}.

RAG pipelines are an instructive partial case, and the verdict depends on where the flow lives. Hardwired in application code, the fixed retrieve-then-generate pipeline \cite{Lewis2020RAG} already fails T1, by the threshold of Section~\ref{sec:definicao}: there is a chain in spirit, but no representation enumerable without running the system. Declared as a framework object, with its two or three nodes and one edge, T1 passes in a degenerate way, and what typically fails is T4, and often T2: the object is not available beyond the run to inspect or optimize, and swapping the generator prompt means editing the same declaration that fixes the flow. The RAG literature itself documents the drift toward richer shapes, adaptive retrieval, routing, feedback loops \cite{Gao2023RAGSurvey}, and graph-structured retrieval couples the two worlds further \cite{Jin2024GraphCoT}. A RAG pipeline becomes prompt graph engineering exactly when its flow is lifted into an explicit, manipulable graph; nothing about retrieval prevents that, and modern adaptive RAG built on graph runtimes is precisely that lift.

Classic workflow engines fail on the opposite side of every neighbor above: they have the graph and lack the prompt. Make, dataflow systems, and scientific workflow managers pass T1 through T4 for their own node kinds \cite{Feldman1979Make,DavisKeller1982Dataflow,Deelman2009Workflows}. What they miss is the object clause of the definition: nodes that are prompt-parameterized model invocations, with the semantics that entails, stochastic outputs, natural-language parameterization, per-call cost and latency, and results whose correctness is not decidable by exit code. That difference is not cosmetic; it changes the engineering. Caching must reason about semantic equivalence, validation must judge text, optimization must rewrite prompts rather than flags. Prompt graph engineering is therefore not workflow engineering applied unchanged to a new node type; it inherits the graph discipline and adds the discipline the new node semantics demands.

The picture that emerges is a genus and a differentia. The genus is the dataflow tradition: explicit, executable, first-class graphs. The differentia is the node: authored, prompt-parameterized model invocations. Thought topologies have graph without authorship; free agent conversations have prompts without structure; classic engines have structure without prompts; single-prompt craft has neither. The concept sits where both hold, and each neighbor's failing condition says which half it lacks.

\section{Applying the Test to Real Systems (RQ4)}
\label{sec:aplicacao}

\begin{figure}[h!]
    \centering
    \includegraphics[width=0.98\linewidth]{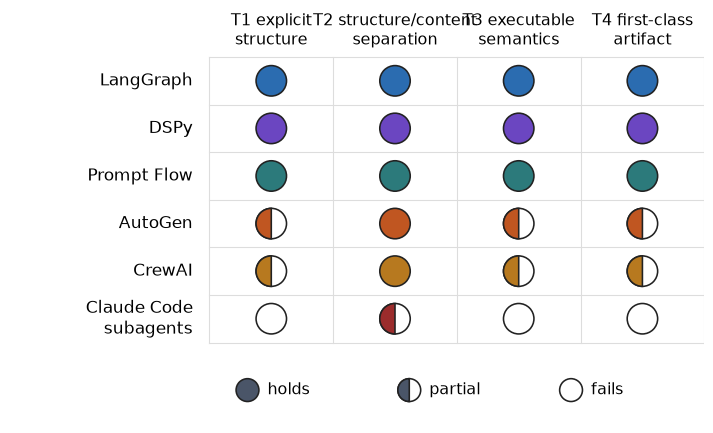}
    \caption{The inclusion and exclusion test applied to six real systems. Filled cells indicate that the condition holds; half-filled cells indicate partial satisfaction (typically one operating mode passes and another fails); empty cells indicate failure. The three fully included systems pass the core cleanly; the multi-agent frameworks split across modes; the coding harness fails the structural conditions, which is exactly the discrimination the test exists to make.}
    \label{fig:componentmap}
\end{figure}

A definition that only works on paper is a slogan. This section applies T1 through T4 to six real systems, chosen to span the practice: two orchestration frameworks born graph-first (LangGraph and Microsoft Prompt Flow), one prompt programming framework (DSPy), two multi-agent frameworks (AutoGen and CrewAI), and one coding harness with subagent delegation (Claude Code). For the framework systems we rely on the primary documentation, cited in footnotes, and on peer-reviewed or preprint literature with persistent identifiers where it exists; the classification is ours, and its criteria are the verifiable ones stated in Section~\ref{sec:definicao}. One scope note is owed up front: these are living systems, so the classification is a dated snapshot, based on documentation accessed in July 2026, and cells of Figure~\ref{fig:componentmap} may move as the frameworks evolve; what does not move is the test that produced them. Figure~\ref{fig:componentmap} summarizes the result; Table~\ref{tab:aplicacao} gives the verdicts.

LangGraph passes the four tests cleanly. The StateGraph is an explicit object built from named nodes and edges, including conditional edges (T1); nodes are functions or prompt-bearing runnables attached to the structure, so either side changes independently (T2); the runtime executes the graph over a shared typed state, with cycles, branching, interrupts, and checkpointing (T3); and the compiled graph is serializable, visualizable, and consumed by external tooling for tracing and evaluation (T4).\footnote{LangGraph documentation: \url{https://langchain-ai.github.io/langgraph/}. Tracing and evaluation via LangSmith: \url{https://docs.smith.langchain.com/}. Accessed July 2026.} Its strength among the six is G3: state, cycles, and human-in-the-loop interrupts are native, which is exactly what fixed DAG tools lack.

DSPy also passes the four, with a different center of gravity. Modules compose into a program whose structure is enumerable (T1); signatures declare what each node consumes and produces while the compiler generates the prompt text, the most radical separation of structure and content in the sample (T2); execution interprets the composition (T3); and the program is the explicit substrate of optimizers that tune prompts and demonstrations against a metric (T4) \cite{Khattab2022DSP,Khattab2023DSPy}. DSPy shows that the concept does not require a canvas: the graph lives in code, and its G4 is the strongest of the six.\footnote{DSPy documentation: \url{https://dspy.ai}. Accessed July 2026.}

Prompt Flow passes the four in the most literal form. A flow is a YAML-declared DAG of prompt nodes and tool nodes (T1); prompt content lives in template files referenced by the structure (T2); a runtime executes the DAG (T3); and the artifact is rendered visually, versioned as files, and fed to batch evaluation (T4).\footnote{Microsoft Prompt Flow documentation: \url{https://microsoft.github.io/promptflow/}. Accessed July 2026.} Its limits mirror its literalness: without native cycles, feedback loops must be pushed inside nodes, which weakens G3 for reflective patterns \cite{Madaan2023SelfRefine}.

AutoGen splits across its two operating modes, and the test registers the split. In conversational mode, agents exchange messages and the interaction shape emerges turn by turn \cite{Wu2023AutoGen}; T1 fails, and what remains after a run is a transcript, not an artifact, so T4 fails too. In GraphFlow mode, the framework reifies the flow as an explicit directed graph over agents with conditional edges, and T1, T3, and substantially T4 pass.\footnote{AutoGen GraphFlow documentation: \url{https://microsoft.github.io/autogen/stable/user-guide/agentchat-user-guide/graph-flow.html}. Accessed July 2026.} In both modes, prompt content (system messages and agent descriptions) is authored and separate from whichever structure, emergent or reified, composes the agents, which satisfies T2. The verdict is partial by construction: AutoGen contains prompt graph engineering as one of its modes, and the emergent mode shows precisely what the concept excludes. StateFlow makes the same movement inside the same ecosystem by modeling task flows as state machines \cite{Wu2024StateFlow}.

CrewAI shows the same duality with different emphasis. Crews compose role-defined agents whose task delegation is partly emergent; the structure of a crew run is not enumerable beforehand, so T1 is at best partial. Flows, the second abstraction, are explicit event-driven compositions with routing and state, and pass T1 and T3.\footnote{CrewAI documentation, Crews and Flows: \url{https://docs.crewai.com/}. Accessed July 2026.} Prompt content (role, goal, backstory, task descriptions) is authored separately from either structure, satisfying T2. T4 is partial: flows are code artifacts and can be versioned, but the crew-level interaction has no graph object to inspect or optimize.

Claude Code subagents are the deliberate counterexample, and the test excludes them for reasons worth stating. The harness lets an orchestrating agent delegate work to authored subagents, each a prompt-parameterized unit; the nodes, in our vocabulary, exist and satisfy the spirit of T2, since subagent definitions are files authored independently of any flow.\footnote{Claude Code subagents documentation: \url{https://code.claude.com/docs/en/sub-agents}. Accessed July 2026.} But which subagent runs, when, and feeding what to what is decided by the orchestrator model at runtime, turn by turn; no representation enumerates the flow beforehand (T1 fails), the runtime executes tool calls rather than a graph (T3 in the graph sense fails), and what persists is a transcript (T4 fails). One scope caveat matters here: the test targets subagent delegation, the mechanism by which the harness composes prompt-bearing units; the harness also exposes authored surfaces, such as commands and hooks, but these automate points of the working loop without reifying the delegation flow into a graph. The delegation topology is real but emergent, closer to the free conversations of multi-agent systems \cite{Wang2024SurveyAgents} than to an authored graph; coding agents in general share this profile, with the agent-computer interface, not a flow artifact, as their structural core \cite{Yang2024SWEagent}. The exclusion is not a demerit: a harness solves a different problem. It merely is not prompt graph engineering, and a definition that included it would have dissolved the concept.

\begin{table}[h!]
\centering
\small
\begin{tabularx}{\textwidth}{|l|c|c|c|c|Y|}
\hline
\textbf{System} & \textbf{T1} & \textbf{T2} & \textbf{T3} & \textbf{T4} & \textbf{Verdict} \\
\hline
LangGraph & yes & yes & yes & yes & Included; strongest in executable semantics (G3) \\
\hline
DSPy & yes & yes & yes & yes & Included; strongest in artifact status and optimization (G4) \\
\hline
Prompt Flow & yes & yes & yes & yes & Included; most literal DAG, weak on cycles \\
\hline
AutoGen & partial & yes & partial & partial & Included in GraphFlow mode; excluded in emergent conversation \\
\hline
CrewAI & partial & yes & partial & partial & Included via Flows; crew-level delegation remains emergent \\
\hline
Claude Code subagents & no & partial & no & no & Excluded; authored nodes, but flow is emergent at runtime \\
\hline
\end{tabularx}
\caption{Verdicts of the inclusion and exclusion test for the six systems analyzed. Partial entries typically indicate systems with more than one operating mode, where one mode passes the condition and another does not.}
\label{tab:aplicacao}
\end{table}

Beyond the six, the research frontier passes the test maximally and confirms the definition's direction of travel. GPTSwarm represents agent societies as computational graphs whose edges are optimized directly \cite{Zhuge2024GPTSwarm}; AFlow searches the space of workflow graphs with tree search \cite{Zhang2024AFlow}; ADAS searches agent designs expressed as code \cite{Hu2024ADAS}; and LLMCompiler builds the dependency DAG dynamically and schedules it \cite{Kim2023LLMCompiler}. All four treat the graph as the object of study, not as plumbing. At the opposite extreme, the everyday counterexamples fail exactly where expected: a script that concatenates two API calls fails T1 and T4; an architecture diagram in a survey fails T3; a single elaborate prompt, however sophisticated its internal reasoning \cite{Wei2022CoT}, fails T1. The test sorts the field into the same regions a practitioner's intuition does, which is what an operational definition is for.

The limits of this exercise should be stated as plainly as its results. The classification was performed by a single analyst; T1 through T4 have verifiable criteria precisely to make the exercise repeatable, but no second rater has yet repeated it, and inter-rater agreement is the obvious first empirical follow-up. The evidence base for the product systems is grey literature, which moves faster than any article; the snapshot caveat above is a mitigation, not a cure. And the sample of six systems, chosen to span the practice, remains a sample: applying the test to serving-layer systems, visual builders, and enterprise orchestrators may stress conditions, especially the T3 threshold, in ways these six do not.

\section{Research Agenda (RQ5)}
\label{sec:agenda}

\begin{figure}[h!]
    \centering
    \includegraphics[width=0.98\linewidth]{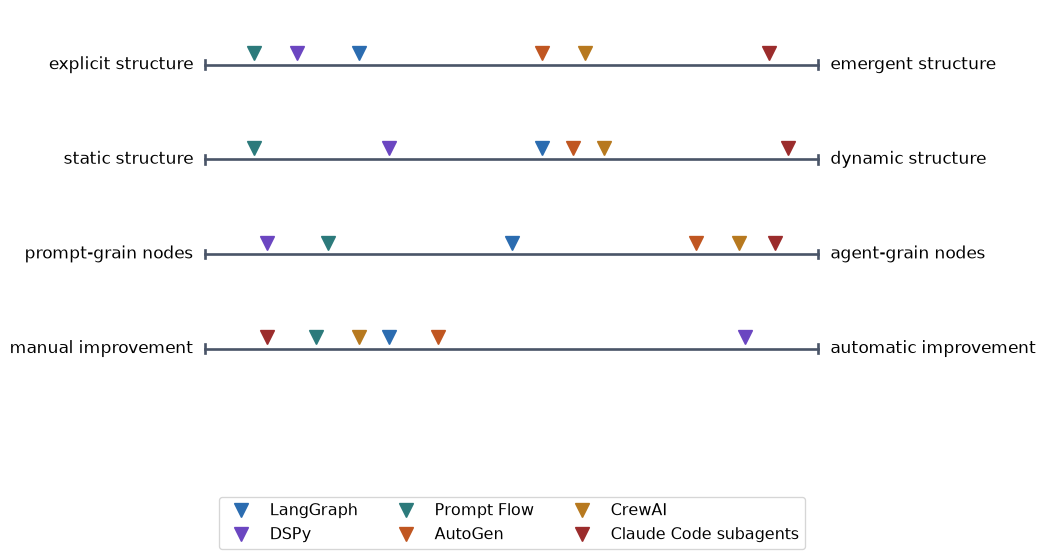}
    \caption{The six analyzed systems positioned on the four design tension axes: explicit versus emergent structure, static versus dynamic structure, prompt-grain versus agent-grain nodes, and manual versus automatic improvement. The positions follow the classification in Section~\ref{sec:aplicacao}; systems with two operating modes appear at the position of their graph-reifying mode.}
    \label{fig:designaxes}
\end{figure}

The application of the test did more than classify; it exposed where real systems disagree. Those disagreements are not noise. They are open design questions, and we organize the agenda along four tension axes, summarized in Figure~\ref{fig:designaxes}.

The first axis runs from explicit to emergent structure. The three graph-first systems author the flow; the multi-agent frameworks let it emerge and reify it only in dedicated modes; the coding harness leaves it emergent entirely. Emergence buys adaptivity: an orchestrator that decides delegation at runtime handles tasks its author never anticipated \cite{Wang2024SurveyAgents,Xi2023RisePotential}. Explicitness buys everything Section~\ref{sec:definicao} enumerated: inspection, verification, optimization. The open question is whether the trade is fundamental or technological: can a system record the emergent flow, lift it into an explicit graph, and replay or refine it, making emergence a discovery mode for structures that then become artifacts? The dynamic DAG construction of LLMCompiler suggests the two ends can meet \cite{Kim2023LLMCompiler}, but, to our knowledge, no system today closes the loop from trace to versioned, optimized graph.

The second axis runs from static to dynamic structure. Prompt Flow fixes the DAG before execution; LangGraph routes conditionally over a fixed node set; LLMCompiler builds the graph per task instance \cite{Kim2023LLMCompiler}; and the thought topologies rebuild it per problem \cite{Yao2023ToT,Besta2024GoT}. More dynamism means more expressiveness and less static knowledge. The interesting middle is barely explored: graphs with static skeletons and dynamically instantiated regions, and type systems that can say something useful about a graph whose shape is partly decided at runtime. The skeleton-first decomposition strategies hint at what such contracts could look like \cite{Ning2023SoT,Khot2023Decomposed}.

The third axis concerns node granularity: prompt-parameterized invocations at one end, full agents with goals, memory, and tools at the other. Fine grain gives analyzable dataflow, as in compiled pipelines \cite{Khattab2023DSPy}; coarse grain gives encapsulation and role clarity, as in agent societies \cite{Li2023CAMEL,Hong2023MetaGPT,Qian2023ChatDev}. Today the choice is forced by the framework rather than made by the engineer, and the design patterns literature records the same split without resolving it \cite{Liu2024DesignPatterns,Guo2024MultiAgentSurvey,Masterman2024Landscape}. The open problem is composition across grains: graphs whose nodes are themselves graphs, with semantics for state, cost, and failure that survive the nesting, the direction the flows abstraction points toward \cite{Josifoski2023Flows}.

The fourth axis runs from manual to automatic improvement. Hand-tuned graphs occupy one end; at the other, prompt optimization \cite{Zhou2023APE,Fernando2023Promptbreeder}, textual gradients propagated through compositions \cite{Yuksekgonul2024TextGrad}, compiled pipelines \cite{Khattab2023DSPy}, and searches over the structures themselves \cite{Hu2024ADAS,Zhang2024AFlow,Zhuge2024GPTSwarm}. Automation presupposes G4 and rewards it; that is the deepest argument for the discipline this article defines. But automated search over graphs whose nodes are stochastic and expensive raises questions the classic AutoML literature never faced: evaluation noise on every fitness call, cost ceilings that bound the search budget, and the risk that an optimizer exploits benchmark idiosyncrasies rather than improving the structure.

Across the four axes, three transversal problems deserve statement. Verification: which properties of a prompt graph can be checked statically, type compatibility along edges, termination of cycles with exit conditions, cost and latency bounds, and which require semantic judgment of text, where verification itself may need model-bearing nodes \cite{Madaan2023SelfRefine,Shinn2023Reflexion}? Context discipline: decomposition into nodes is also a context management strategy, since each node sees a curated window rather than an accumulated history whose middle the model loses \cite{Liu2024LostMiddle}; the interaction between graph shape and per-node context quality is unmeasured. And equivalence: when are two prompt graphs the same program, what refactorings preserve behavior in distribution, and can retrieval-heavy and reasoning-heavy shapes be compared on common ground \cite{Gao2023RAGSurvey,Jin2024GraphCoT}? Tool-use research keeps adding node vocabularies \cite{Schick2023Toolformer,Yao2023ReAct}, and each new vocabulary re-poses all three problems.

\section{Conclusion}
\label{sec:conclusao}

Practice ran ahead of vocabulary. Engineers already compose prompts into routed, parallel, cyclic structures every day, and frameworks already expose those structures as their central abstraction, yet the discipline had no definition that says what it is and what it is not. We reversed that order. We reconstructed the genealogy of the prompt graph, from dataflow models and build systems, through chaining and the thought topologies, to graphs that are compiled and searched as objects. The thread is single: the graph earns its keep whenever computation must be seen, divided, and controlled, and prompts became computation worth all three.

On that base we defined prompt graph engineering by four conditions, explicit structure, separation of structure and prompt content, executable semantics, and the graph as a first-class artifact, and turned the definition into an inclusion and exclusion test. The test drew a boundary that neighbors fail for nameable reasons: classic prompt engineering has no structure to represent; thought topologies have graphs the model, not the engineer, authors; free agent conversations leave no artifact; classic workflow engines have the graph but not the prompt. Applied to six real systems, the test included the graph-first frameworks, split the multi-agent frameworks along their own operating modes, and excluded a coding harness whose delegation is real but emergent; it sorts the field the way practitioner intuition does, with reasons attached.

At bottom, the contribution is conceptual hygiene with an engineering payoff. The four conditions are not bureaucracy; each one is the precondition for something practice already wants: inspection needs explicit structure, reuse and compilation need separation, execution needs semantics, and the entire optimization lineage needs the artifact. The definition also opens the question it cannot answer alone: how much of a system's quality lives in the structure rather than in the prompts, and when does lifting an emergent flow into an explicit graph pay for itself? Those are empirical questions, and they need the vocabulary this article fixed before they can be asked cleanly. Knowing what prompt graph engineering is was the necessary step toward measuring, next, what it is worth.

\section*{Declaration on the Use of Generative AI}

The author conducted the research and wrote the manuscript. During the preparation of this study, however, the author used Grammarly tools to improve textual agreement and Claude Opus 4.8 to support text structuring and translation into English. After using these tools/services, the author reviewed and edited the content as needed and takes full responsibility for the content of the publication.

\bibliographystyle{plain}
\bibliography{refs}

\end{document}